\documentclass[review]{elsarticle}

\usepackage{lineno,hyperref}
\usepackage{float,booktabs,multirow}
\usepackage{xcolor}
\usepackage{subcaption,caption}
\usepackage{hyperref}
\usepackage[lmargin=3.00cm,rmargin=3.00cm,tmargin=3.0cm,bmargin=3.0cm]{geometry}
\usepackage{blindtext}
\usepackage{enumitem}
\modulolinenumbers[5]

\journal{NeuroImage}

\begin{document}

\begin{frontmatter}

\title{An automatic pipeline for atlas-based fetal and neonatal brain segmentation and analysis}

\author[1]{Andrea Urru}
\author[2,3]{Ayako Nakaki}
\author[4]{Oualid Benkarim}
\author[2]{Francesca Crovetto}
\author[2]{Laura Segales}
\author[1]{Valentin Comte}
\author[2,3]{Nadine Hahner}
\author[2,3,5]{Elisenda Eixarch}
\author[2,3,5]{Eduard Gratac\'os}
\author[2,3,5]{F\`atima Crispi}
\author[1]{Gemma Piella}
\author[1,6]{Miguel A. Gonz\'alez Ballester\corref{cor1}}
\cortext[cor1]{Corresponding author: ma.gonzalez@upf.edu}
\address[1]{BCN MedTech, Department of Information and Communication Technologies, Universitat Pompeu Fabra, Barcelona, Spain}
\address[2]{BCNatal | Fetal Medicine Research Center (Hospital Clínic and Hospital Sant Joan de Déu), University of Barcelona, Barcelona, Spain}
\address[3]{Institut d'Investigacions Biomèdiques August Pi i Sunyer (IDIBAPS), Barcelona, Spain}
\address[4]{McConnell Brain Imaging Centre, Montreal Neurological Institute and Hospital, McGill University, Montreal, Quebec, Canada}
\address[5]{Centre for Biomedical Research on Rare Diseases (CIBERER), Barcelona, Spain}
\address[6]{ICREA, Barcelona, Spain}
%%\fntext[myfootnote]{}

%% \author[mymainaddress,mysecondaryaddress]{Elsevier Inc}
%% \ead[url]{www.elsevier.com}

%% \author[mysecondaryaddress]{Global Customer Service\corref{mycorrespondingauthor}}
%% \cortext[mycorrespondingauthor]{Corresponding author}
%% \ead{support@elsevier.com}

%% \address[mymainaddress]{1600 John F Kennedy Boulevard, Philadelphia}
%% \address[mysecondaryaddress]{360 Park Avenue South, New York}

\begin{abstract}
The automatic segmentation of perinatal brain structures in magnetic resonance imaging (MRI) is of utmost importance for the study of brain growth and related complications. While different methods exist for adult and pediatric MRI data, there is a lack for automatic tools for the analysis of perinatal imaging. In this work, a new pipeline for fetal and neonatal segmentation has been developed. We also report the creation of two new fetal atlases, and their use within the pipeline for atlas-based segmentation, based on novel registration methods. The pipeline is also able to extract cortical and pial surfaces and compute features, such as curvature, thickness, sulcal depth, and local gyrification index. Results show that the introduction of the new templates together with our segmentation strategy leads to accurate results when compared to expert annotations, as well as better performances when compared to a reference pipeline (developing Human Connectome Project (dHCP)), for both early and late-onset fetal brains.
\end{abstract}


\end{frontmatter}

\section{Introduction}
\label{ch:Introduction}

Perinatal brain imaging has drawn increasing attention for clinical purposes, as many diseases can be identified already at fetal or neonatal stage by monitoring the development of different brain tissues and structures \cite{benkarim2018cortical, wright2000meta, 
%%lyoo1996corpus, 
jackson2011ventricular, Benkarim2020, Hahner2019}. Early diagnosis of abnormal development has thus revealed to be of utmost importance for improved treatments and follow-up. Advances in imaging techniques such as Magnetic Resonance Imaging (MRI) made it possible to obtain highly detailed 3D images of the brain. Compared to the adult brain, however, perinatal brain MRI encounters a number of non-trivial challenges, among which increased motion artifacts, especially in fetal MRI, temporal variations in tissue intensities, low contrast-to-noise ratio and the high growth rate of the brain in this period, which makes monitoring its development through time more difficult. Although being largely investigated, this topic remains an open challenge in image analysis \cite{benkarim2017toward}. 

This work introduces the first segmentation pipeline aimed at seamlessly addressing the analysis of the whole perinatal period, both pre- and post-natal. Over the years, several algorithms have been proposed for fetal or neonatal image analysis (see \cite{benkarim2017toward, fe03cf7f6f444777b79732505952b3bb} for reviews); however, the methods developed so far present different limitations. In particular, they aim to label the whole brain, different tissues or local structures and substructures in fetal or neonatal brains. Nevertheless, there is a lack for an end-to-end tool for an automated analysis of both fetal and neonatal acquisitions, which would allow for a more advanced and complete analysis of the brain development, combining an automatic tissue segmentation to a regional, structural segmentation. 

Related works include the dHCP pipeline \cite{makro2018}, which aims to build a neonatal brain connectome starting from the acquired MRI, but does not tackle the challenges of fetal acquisition, segmentation and reconstruction. In \cite{payette2020efficient} an annotated dataset composed of 50 fetal brains is used to compute a tissue segmentation by supervised learning techniques and multi-atlas segmentation. It focuses on prenatal segmentation, but it does not integrate the brain super-resolution reconstruction (SRR) \cite{ebner2020} and skull extraction as pre-processing steps. Moreover, only the tissue segmentation is computed; no structural or regional information and no cortical surface mesh estimation are generated.

In this work, a newly developed framework for fetal and neonatal segmentation and analysis is proposed. The pipeline is built upon the existing dHCP pipeline \cite{makro2018} for neonatal segmentation, and it introduces:
\begin{enumerate}
  \item A new fetal multi-tissue segmentation based on the brain spatio-temporal probabilistic atlas \cite{630093}. \
  \item A newly developed multi-subject fetal atlas, built from new data acquired at our institution, and its application for tissue and structural segmentation. This atlas, shared to the community, is in itself an important contribution to the field.\
  \item For fetal acquisitions, highly affected by motion-related noise, a super-resolution reconstruction, using a 2D U-net based skull stripping method \cite{salehi2018} and the NiftyMIC SRR framework \cite{ebner2020}, added as a pre-processing step to the pipeline. \
  \item Structural and multi-tissue segmentation, computed through a multi-channel, multi-resolution non-linear registration to the developed templates, using the ANTs toolbox \cite{avants2009advanced}.
  \item The pipeline's source code and atlases, as well as its installation and usage guidelines are publicly available at \href{https://github.com/urrand/perinatal-pipeline}{https://github.com/urrand/perinatal-pipeline}.
\end{enumerate}

We next provide background information about brain segmentation methods, as well as about fetal and neonatal atlases. Then, we describe the overall structure of the pipeline and elaborate on its different elements. We present its use on a clinical dataset composed of a total of 150 fetal subjects and 30 neonates. The method shows good performance, both in terms of visual inspection and through quantitative experiments reporting the Dice score with respect to ground truth segmentations. Furthermore, we present experiments comparing our method with the original dHCP pipeline \cite{makro2018}, and show increased performance in terms of segmentation accuracy. Finally, discussions and potential directions for future research are described.

\section{Background}
\label{ch:Background}

\subsection{Brain segmentation techniques}
Following the criteria of \cite{fe03cf7f6f444777b79732505952b3bb}, the segmentation techniques proposed so far for brain MR images can be divided in five main groups based on the methodology they adopt: unsupervised, parametric, classification-based, deformable models, and atlas fusion methods. 

Unsupervised techniques often leverage traditional segmentation algorithms (e.g. thresholding, region growing, morphological operations) and are not based on previously labelled training data. They are typically highly sensitive to noise. They can be used to correct for partial volume voxels, or as a tool to compute tissue priors \cite{xue2007automatic, makro2018}. 
Parametric techniques propose a segmentation by fitting a model, typically a Gaussian Mixture Model (GMM), to the data using an Expectation-Maximization (EM) approach \cite{van1999automated, GonzalezBallester2000, xue2007automatic, makropoulos2012automatic}. Classification techniques use a classifier trained on a given dataset to learn how to assign a certain label to voxels or groups of voxels, based on selected features. Previous works in literature used k-NN, decision forests and, lately, convolutional neural networks (CNNs) for both brain extraction \cite{keraudren2014automated, kainz2014fast, rajchl2016deepcut, serag2016accurate} and tissue classification \cite{moeskops2016automatic, sanroma2016building}. 
Deformable models aim to segment an object via deformation of a surface based on physical internal and external energies that move locally the surface depending on the properties of the image. The external energy is usually driven by the image properties, while the internal force is used to smooth and constrain the resulting deformation. An example of deformable model-based technique is used for brain extraction and constitutes the core of the Brain Extraction Tool (BET) \cite{smith2002fast}. 

Finally, another group of techniques use atlases \cite{OISHI20118, serag2012multi, 630093, kuklisova2011dynamic, serag2012construction, schuh2014construction, habas2010spatiotemporal, Sanroma2018}, which provide a representation of the average tissue distribution and its variability in the population. The segmentation is obtained by registering the target image to the atlas. Once the alignment has been performed, label information is transferred from the atlas to the target image. Multi-subject atlases are composed of different subjects instead of a single one, representing the population. In this case, the target image is registered to each subject composing the atlas and once the alignment is performed, label information is transferred from the different subjects through a weighted average. Multi-subject atlas fusion can be used both for brain extraction \cite{tourbier2015automatic} and tissue or structural segmentation \cite{gousias2013magnetic}.

\subsection{Fetal and neonatal atlases}
In the context of fetal and neonatal brain analysis, construction of corresponding atlases has been of paramount importance because they have been used not only to represent the average growth of the brain and its variation across the population, but also to automatically segment single subjects at different developmental stages \cite{serag2012multi, 630093}. Currently, perinatal brain atlases can be classified in two main categories: probabilistic atlases and multi-subject atlases. Both approaches have been introduced to overcome the limitations of single-subject atlases \cite{OISHI20118}, which are based on a subject taken as a reference for the population, regardless of the inherent variability in shape, size and growth rate of the brain at this stage of development. 

Probabilistic atlases present a brain resulting from the average of the intensity images and segmentations of all the subjects in the cohort. In several works present in literature, a spatio-temporal probabilistic atlas is built using non-parametric kernel regression \cite{kuklisova2011dynamic, serag2012construction, schuh2014construction, habas2010spatiotemporal}. Although these works have a similar approach to compute a high-definition average of the subjects in the cohort for each time-point, they differ in the registration technique used to align the subjects: either a simple affine registration \cite{kuklisova2011dynamic}, a non-rigid pairwise registration between all the subjects \cite{serag2012construction, schuh2014construction}, or a groupwise registration \cite{habas2010spatiotemporal}. Other works aimed to enhance these atlases, adding new labelled structures to the ones initially proposed \cite{makropoulos2016regional}. To obtain the segmentation for a new subject, this is registered to the template and the corresponding segmentation is computed using the resulting transformation. Since the template is smoothed out by the average, the registration is more accurate and less challenging than a registration between two individual subjects. To take into account variability across the population, probability maps for different tissues can be built, indicating for each voxel in the template its probability of belonging to each structure, or tissue. 

Multi-subject atlases are another solution to represent, up to a certain extent, variability across the population; in this case, a new subject is registered to each one of the volumes in the atlas independently, and transformations are averaged out to obtain the final segmentation for the subject. Most atlases present in literature are composed of healthy subjects at a neonatal stage of development \cite{gousias2012magnetic, alexander2017new}, although a dataset composed by both normal and pathological fetal brains \cite{Payette_2021} has been proposed and used in different recent research works \cite{payette2020efficient, dedumast2020segmentation}. 

Limitations in this field are related to the different segmentations proposed by each atlas, which makes comparison between pre- and post-natal segmentations difficult. Well established atlases such as \cite{kuklisova2011dynamic, SERAG20122255, 630093} present different segmentations of the same brain tissues. As regards structural segmentation, although the atlases in \cite{gousias2012magnetic, makropoulos2016regional} propose a regional labeling for neonatal brain, there is not an equivalent multi-subject atlas for fetal brains.

\section{Methodology}
\label{ch:Methods}

\begin{figure}[H]
\centering
\includegraphics[width=\linewidth]{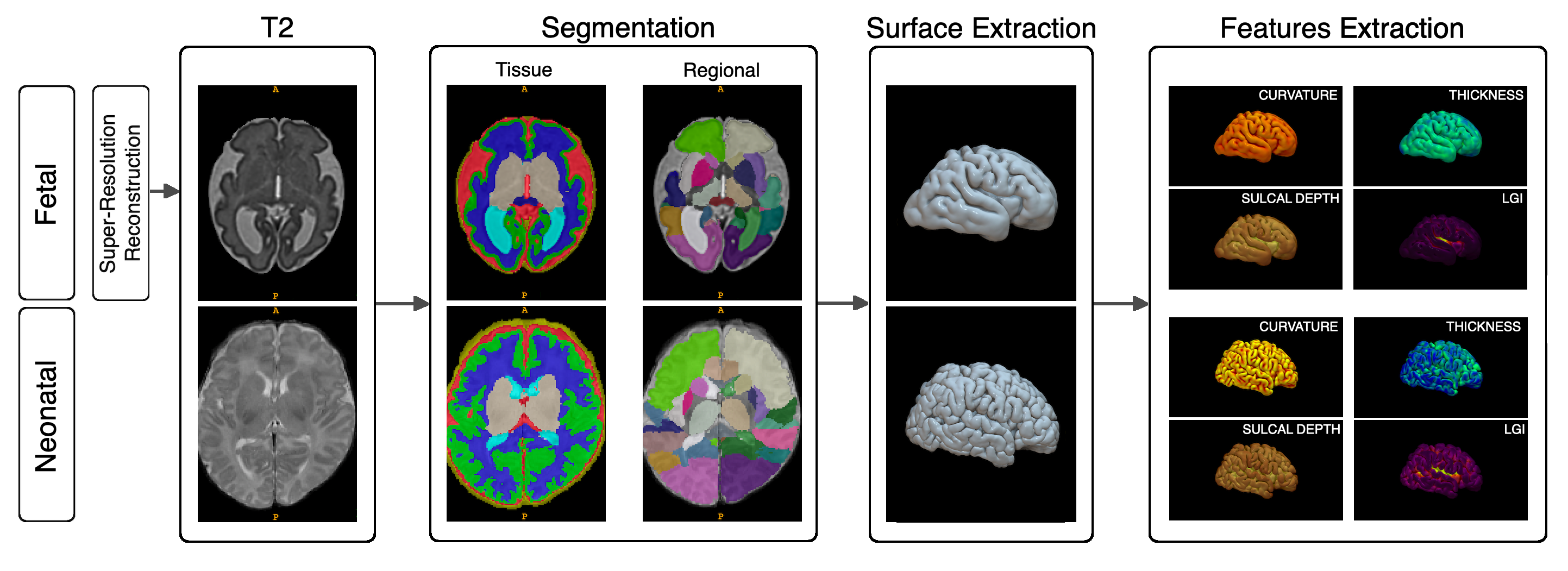}
\caption{Perinatal pipeline workflow.}
\label{fig:SegPipeline}
\end{figure}

The workflow of the proposed pipeline is summarised in Fig. \ref{fig:SegPipeline}. It takes as an input the raw data of the image to be segmented (fetal or neonatal). For fetal acquisitions, SRR is applied, in order to combine the different imaging stacks into one high-resolution image volume. Skull extraction and the N4 bias correction are then applied. The tissue priors for gray matter and ventricles are computed registering the obtained T2-weighted reconstructed image to the temporal templates, and they are used with the T2 image in a 3-channel registration to the structural multi-subject atlases, to obtain the tissue and structural labels. Finally, starting from the pial and white matter boundaries, the cortical surface is extracted.

All these steps are described in more detail in the subsections below. First, we will describe the imaging datasets and the atlases built and used by our pipeline.

\subsection{Dataset acquisition and reconstruction}
\subsubsection{Fetal Dataset}
The subjects used for the fetal brain MRI dataset of this study were randomly selected pregnant participants who participated in a large randomized clinical trial \cite{crovetto}. All fetuses included in this study did not have any major malformation. All participants provided written informed consent on the day of recruitment for the fetal MRI. The protocol was approved by the Ethics Committee of Hospital Clínic Barcelona, Spain (HCB/2020/0267).  A total of 200 participants underwent fetal MRI between 32 weeks and 39 weeks of gestational (mean 36.5 ± 1.0 weeks SD), with the proportion of 105 (52\%) male fetuses.

Single-shot fast spin-echo T2-weighted was performed on two 3.0 T MR scanners (Philips Ingenia and SIEMENS MAGNETOM Vida) in two hospitals (Hospital San Joan de Déu and Hospital Clínic), respectively, using a body array radio-frequency coil without sedation. The parameters used for each machine were as follows: 1) Philips: repetition time 1570 ms, echo time 150 ms, slice thickness 3 mm, field of view 290x250 mm, voxel spacing 0.7x0.7x3.0 mm, no interslice-slice gap; 2) Siemens: repetition time 1390 ms, echo time 160 ms, slice thickness 3 mm, field of view 230x230 mm, voxel spacing 1.2x1.2x3.0 mm, no interslice-slice gap.
Three orthogonal planes, oriented along the axis of the fetal brainstem, obtaining 2-loops of axial, 2-loops of coronal and 2-loops of sagittal single shoot slices were obtained for each subject. Among the 200 participants, 91 underwent MRI in Hospital San Joan de Déu between 32 to 39 weeks of gestation (mean 36.5 ± 1.1 SD), and 109 in Hospital Clínic between 35 to 39 weeks of gestation (mean 36.3 ± 0.7 SD). 

\subsubsection{Neonatal Dataset}
The 30 subjects used for the neonatal dataset of this study were part of a prospective study about fetal ventriculomegaly (REF). Babies were scanned at XX post-menstrual age (PMA), during natural sleep using a TIM TRIO 3.0 T whole body MR scanner (Siemens, Germany). The study protocol and the recruitment and scanning procedures were approved by the Institutional Ethics Committee, and written informed consent was obtained from the parents of each child to participate in the research studies (HCB/2014/0484). T2 weighted images were obtained with the following parameters: 2-mm slice thickness with 2 mm interslice gap, in-plane acquisition matrix of 256×256, FoV=160X241 mm2, which resulted in a voxel dimension of 0.625×0.625×2 mm3, TR=5980 ms and TE=91 ms. All acquired MRI images were visually inspected for apparent or aberrant artifacts and brain anomalies and subjects excluded accordingly.

\subsection{Atlases}

In order to cover the complete perinatal period, both neonatal and fetal atlases are needed. As our pipeline is based on the dHCP pipeline \cite{makro2018}, we use their atlases for neonatal data. For fetal data, however, new atlases need to be developed. In the following subsections we describe how these fetal atlases have been built, both for the temporal template and the multi-subject atlas (based on a new in-house fetal dataset).

\subsubsection{Fetal temporal template}

\begin{figure}[H]
  \begin{subfigure}[b]{0.49\textwidth}
    \includegraphics[width=\textwidth]{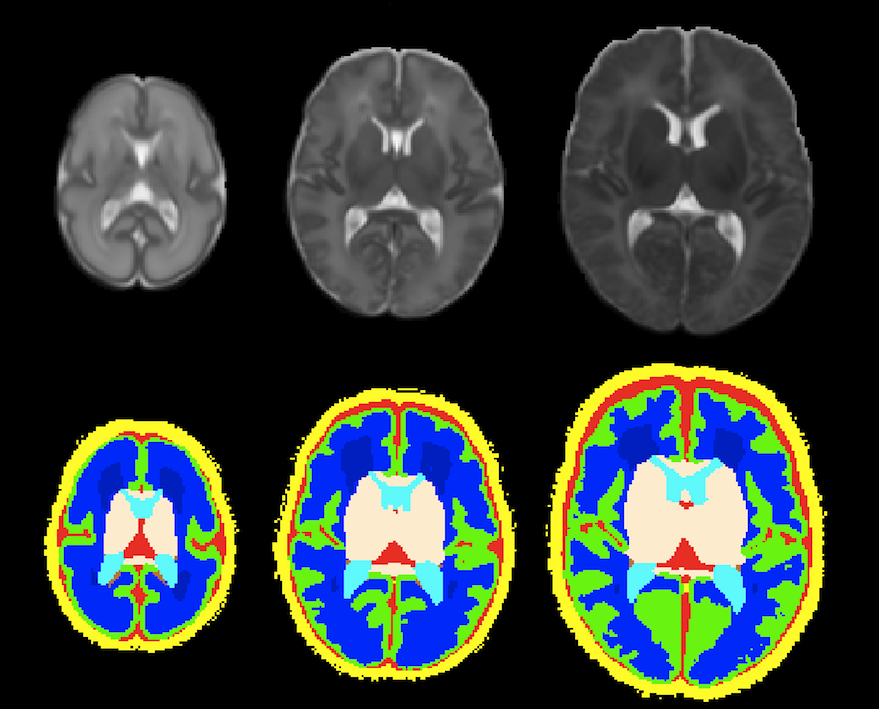}
    \caption{}
    \label{fig:1a}
  \end{subfigure}%
  \hfill
  \begin{subfigure}[b]{0.49\textwidth}
    \includegraphics[width=\textwidth]{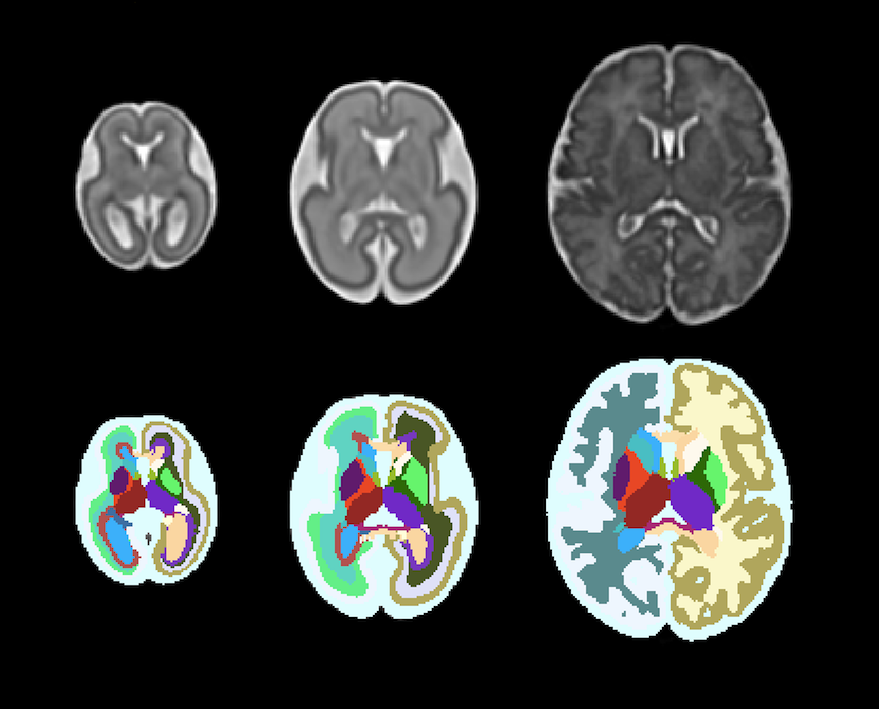}
    \caption{}
    \label{fig:1b}
  \end{subfigure}
\caption{Neonatal and Fetal temporal templates with corresponding original segmentations. (a) Neonatal template at 28, 36 and 44 weeks of gestation respectively. (b) Fetal template at 19, 29 and 39 weeks of gestation respectively.}
\end{figure}

In the dHCP pipeline, a neonatal temporal template \cite{SERAG20122255} is used to estimate the initial tissue segmentation. This template was constructed using T1 and T2 weighted MR images from 204 premature neonates (no preterm babies had visually obvious pathology). The age range at the time of scan was 26.7 to 44.3 weeks of gestation, with mean and standard deviation of 37.3 ± 4.8 weeks. All subjects were born prematurely, with mean age at birth 29.2 ± 2.7, in a range between 24.1 to 35.3 weeks of gestation. The result is a temporal template covering gestational ages between 28 and 44 weeks, and the segmentation identifies 6 tissues and the background (and additionally, it distinguishes high and low intensity white matter). The atlas and the corresponding segmentation is shown in Figure \ref{fig:1a}.

In this work, an enhanced version of \cite{630093} for tissue segmentation has been added to the pipeline for fetal segmentation. This atlas has originally been built from T2 weighted MR images of 81 healthy fetuses scanned at a gestational age range between 19 and 39 weeks, with mean and standard deviation of 30.1 ± 4.5 weeks; for each week, an average shape is computed. Two types of segmentation are proposed, one identifying the tissues and the other, based on \cite{gousias2012magnetic}, brain structures and regions, for a total of 124 labels. In order to identify the equivalent 7-tissues segmentation, this atlas is registered to the neonatal template: each of the computed average brains composing the fetal template is registered to every average brain in the neonatal one, and the final segmentation is computed as a locally weighted average of the neonatal atlas segmentation, based on the local similarity after the registration. For the registration, the ANTs toolbox \cite{avants2009advanced} has been used combining a rigid, affine and non-rigid registration in a multi-resolution approach, with mutual information selected as similarity metric. A multi-channel registration has been implemented, in which the T2-weighted image, the cortical gray matter and ventricles ground truth segmentations are used to improve the registration accuracy. A few samples from the atlas and the original tissue segmentation are shown in Figure \ref{fig:1b}.

\hfill \break

Once registration is performed for every time-point in the neonatal template with respect to each time-point of the fetal template, the correspondence between fetal and neonatal labels is established by simple majority voting among the registered samples. Simple majority voting assigns a label and a probability for every voxel based on the most frequent class assignment, whereas the voxel is not initially classified in case of ties. Subsequently, a label is assigned to these unclassified voxels based on a weighted majority voting among the 26-connected neighborhood with their probabilities.

\subsubsection{Multi-subject fetal atlas}

In order to compute a structural segmentation for a neonatal subject, in the dHCP pipeline the ALBERTs atlas \cite{gousias2012magnetic,10.1371/journal.pone.0059990} is used. This multi-subject atlas is composed of 20 manually labelled subjects and provides T1 and T2 images accompanied by the label map for each subject. 15 of the 20 subjects were born prematurely at a median age at birth of 29 weeks (range 26-35 weeks). These were scanned at term at a median age of 40 weeks (range 37-43 weeks) and had a median weight of 3.0 kg (range 2.0-4.0 kg) at the time of scan. The remaining 5 subjects were born at term at a median age of 41 weeks (range 39-45 weeks), and had a median weight of 4.0 kg (range 3.0-5.0 kg).
 
In this work, a corresponding dataset of $N=20$ fetal subjects has been selected from our cohort to replicate the function of neonatal ALBERTs template for the fetal cases. As a first step, the $N$ subjects' T2-weighted images have been selected within the cohort to maximise variability in terms of age, brain development and shape. These subjects were scanned at a median age of 36.6 weeks (range 33-38 weeks).

After a pre-processing phase consisting of intensity inhomogeneity correction \cite{tustison2010} and skull extraction \cite{smith2002fast}, each subject is initially segmented using the fetal atlas previously described. In this case, a pairwise intensity-based registration is performed selecting the closest sample in the atlas in terms of gestational age, and the segmentation is computed based on the registration outcome. The resulting tissue segmentation has been refined by an expert and constitutes the ground truth segmentation.

\begin{figure}[H]
\centering
\includegraphics[width=\linewidth]{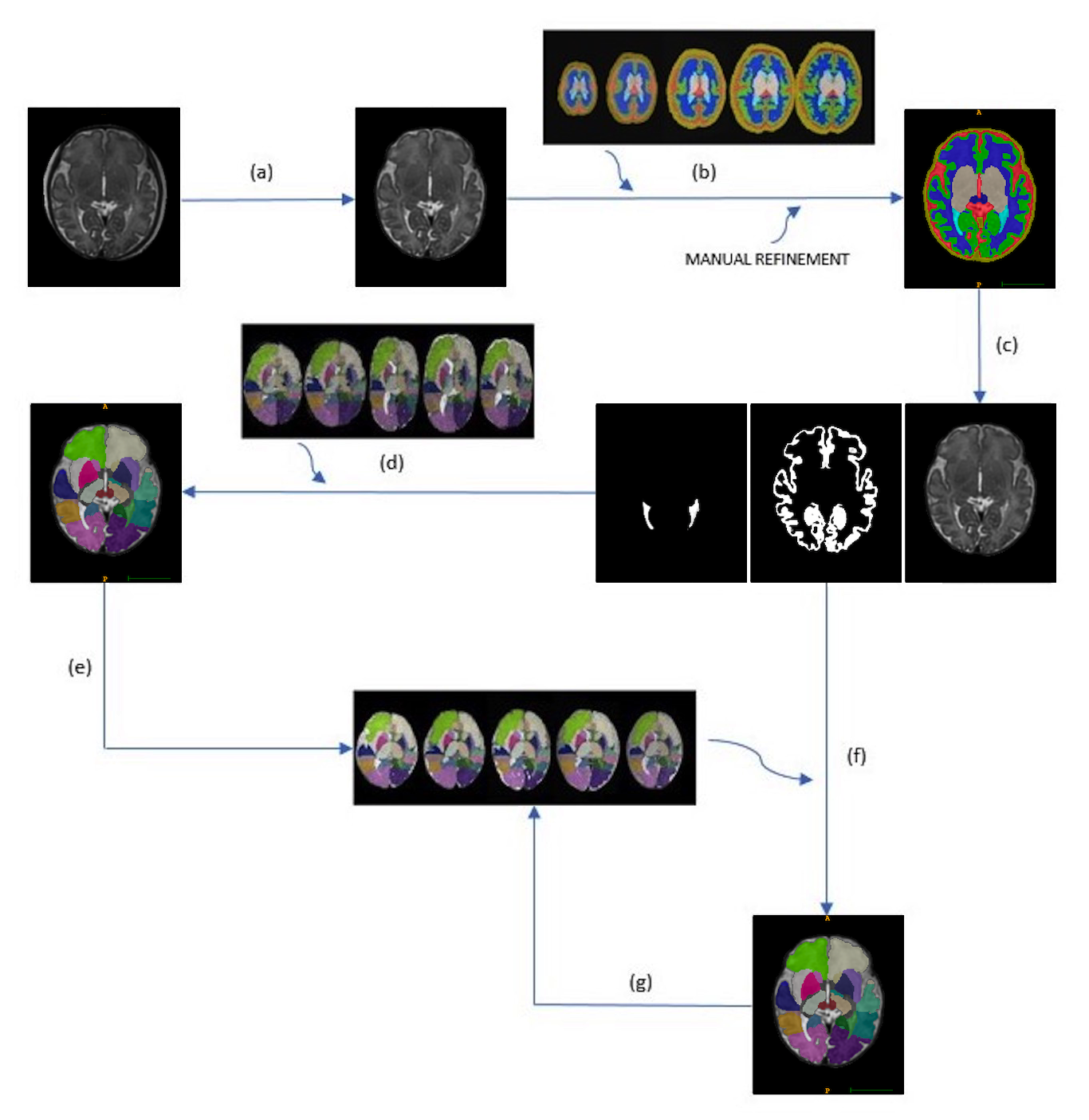}
\caption{The process to generate the multi-subject fetal atlas starts from the reconstructed T2-weighted images. A pre-processing phase (a) consisting of N4 bias correction and brain extraction is performed first, followed by the tissue segmentation (b) using the modified fetal template and a manual refinement. From the obtained segmentation, gray matter and ventricles segmentation are extracted (c). They are used in a three-channel registration with the ALBERTs atlas (d) to obtain a first estimate of their structural segmentation and generate the multi-subject atlas (e). Each subject is then segmented iteratively using the other 19 subjects (f) of the atlas until convergence, and at each step the atlas is updated (g).}
\label{fig:template}
\end{figure}

Subsequently, in order to replicate the function of ALBERTs atlases, the same structural segmentation is needed. To compute it for the selected subjects, they have been segmented using ALBERTs atlases to obtain a first estimate of their structural segmentation, using a three-channel registration with the T2-weighted images and the gray matter and ventricle segmentations. Next, each of the $N$ subjects has been segmented using $N-1$ subjects composing the multi-subject atlas (all the subjects in the dataset but the subject analyzed) to refine their structural segmentation. The resulting structural segmentation for each subject is iteratively refined until convergence. The process is summarized in Figure \ref{fig:template}.

\subsection{Segmentation}
\label{section:Segmentation}

We next describe the full processing pipeline (Fig. \ref{fig:SegPipeline}), which uses the atlases described above to perform segmentation on both fetal and neonatal data. For the case of fetal data, the 3D volume has to be reconstructed first from the different acquisitions along three orthogonal axes. In our pipeline, SRR is implemented \cite{ebner2018, ebner2020} to obtain the 3D image of the fetal brain. Segmentation proceeds with a brain-extraction step, implemented using BET \cite{smith2002fast}, in order to remove the remaining non-brain tissue. Indeed, in neonates the skull, eyes and other structures are present in the image, whereas in fetuses sometimes parts of the skull or even the mother uterus are visible after the reconstruction. BET segments the brain tissue and cerebrospinal fluid (CSF), and removes most of the skull. Once the brain has been extracted, it is rigidly registered to the temporal templates using principal components. The pre-processing phase ends with the intensity inhomogeneity correction using the N4 algorithm \cite{tustison2010}.

\begin{figure}[H]
  \begin{subfigure}[b]{0.49\textwidth}
    \centering
    \includegraphics[height=225bp]{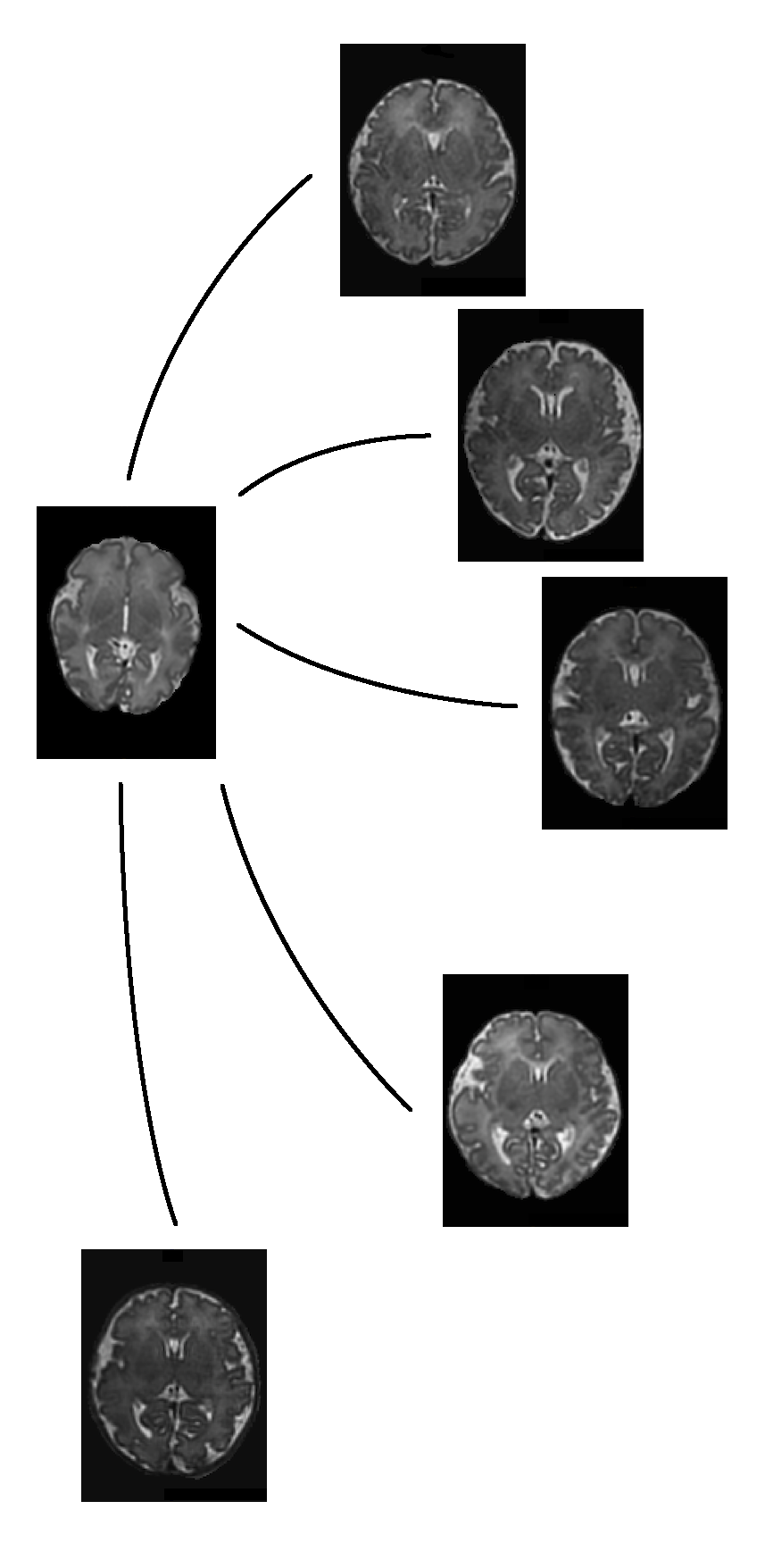}
    \caption{}
    \label{fig:4a}
  \end{subfigure}%
  \hfill
  \begin{subfigure}[b]{0.49\textwidth}
    \centering
    \includegraphics[height=225bp]{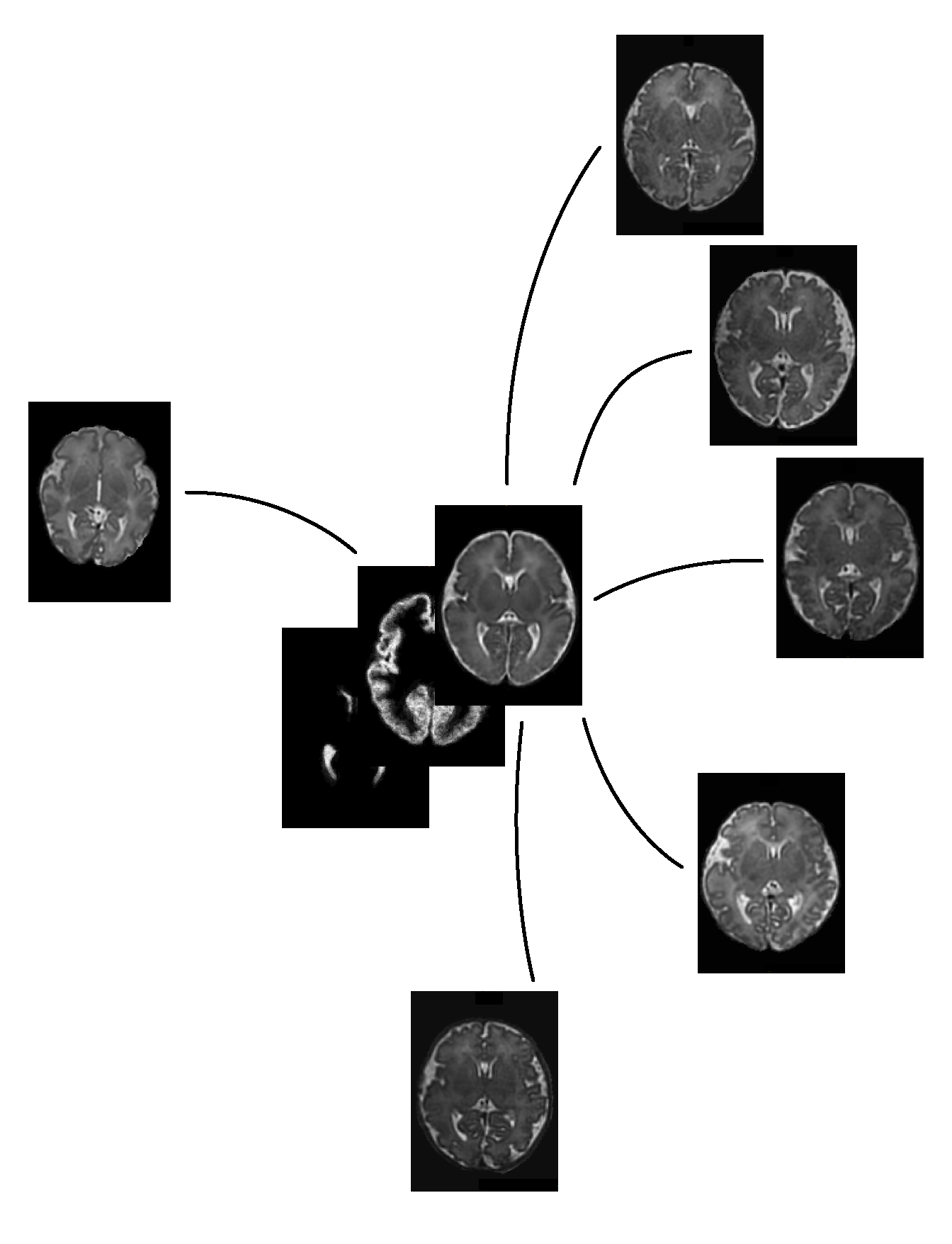}
    \caption{}
    \label{fig:4b}
  \end{subfigure}
\caption{(a) Registration phase as performed in the dHCP pipeline: the subject (left) is registered to each one of the subject composing the multi-subject atlas; (b) Registration phase as performed in the proposed pipeline: the subject is registered to the computed template, and the resulting transformation combined with the previously stored transformations between the template and the subjects composing the multi-subject atlas.}
\label{fig:RegComparison}
\end{figure}

Subsequently, segmentation is performed with an Expectation-Maximization algorithm. With respect to the dHCP pipeline, we have chosen a different approach for registration in order to reduce the required computational load. In particular, in the dHCP pipeline the subject T2-weighted image is registered to each one of the subjects composing the multi-subject atlas (i.e., a total of 20 multi-resolution registrations using the ALBERTs multi-subject atlas). Here, we reduce the amount of required registrations to one. That is done by computing a single template for the multi-subject atlas and by storing the deformation fields from each one of the subjects to the template. The single template is obtained by registering each of the subjects to every other one in the atlas and applying the average deformation field before averaging the resulting warped images, in an iterative approach. When the pipeline is launched, only the transformation between the examined subject and the template will be computed, and the deformation will be combined with the ones already stored. Moreover, a 3-channel registration using the T2-weighted image, gray matter, ventricle probability maps is used, instead of the 2-channel registration implemented in the dHCP pipeline. The comparison between the two approaches is summarised in Figure \ref{fig:RegComparison}.

The probability maps of the different brain tissues are computed using multiple labelled atlases (i.e., ALBERTs atlases in the case of neonates). The atlas labels are transformed and weighted based on their local similarity with respect to the target after registration to merge multiple segmentations into a final weighted structural segmentation, a tissue segmentation and the white and pial boundaries.

\subsection{Surface extraction}

Following segmentation, white-matter and pial surfaces extraction is performed. First, a triangulated surface mesh is fit onto the computed white-matter segmentation boundary. The shape of the mesh is corrected considering possible holes in the segmentation, using the N4 bias-corrected T2-weighted image \cite{schuh2017}. The deformation of the initial surface is driven by a balance of internal and external forces. While the external force seeks to minimise the distance between the mesh vertices and the tissue boundaries, three different internal forces aim to regularise the deformation. They enforce smoothness and avoid self-intersections by introducing a repulsion force and a minimum distance between adjacent vertices and triangles in the mesh. If a movement brings a triangle under the minimum distance from another triangle, the force of its vertices is limited and the movement prevented. Thus, in the first stage, external forces are derived from distances to the tissue segmentation mask, and in the second stage, boundaries are refined using external forces derived from intensity information.
The pial surface is obtained by deforming the white-matter mesh towards the boundary between cortical gray matter (cGM) and CSF, modifying the external force to search for the closest cGM/CSF image edge outside the white-matter mesh. 
White matter and pial surface extraction allow computing features of the cortical surface, in particular surface curvature, cortical thickness, sulcal depth (i.e. the average convexity or concavity of cortical surface points) and local gyrification index (i.e. a measure of the amount of surface buried in the sulci with respect to the one visible on the surface).
More details on the surface extraction process can be found in \cite{makro2018}.

\section{Results}
\label{ch:Results}

\subsection{Temporal template}

The fetal template proposed in \cite{630093} has been segmented according to the neonatal template presented in \cite{SERAG20122255} and originally used in the dHCP pipeline. The neonatal template presented a 7-tissue segmentation, and is used in the dHCP pipeline to have an initial estimation of the tissue segmentation, in particular the grey matter priors, as part of the registration process to obtain the final structural segmentation.

\begin{figure}[H]
\centering
\includegraphics[width=\linewidth]{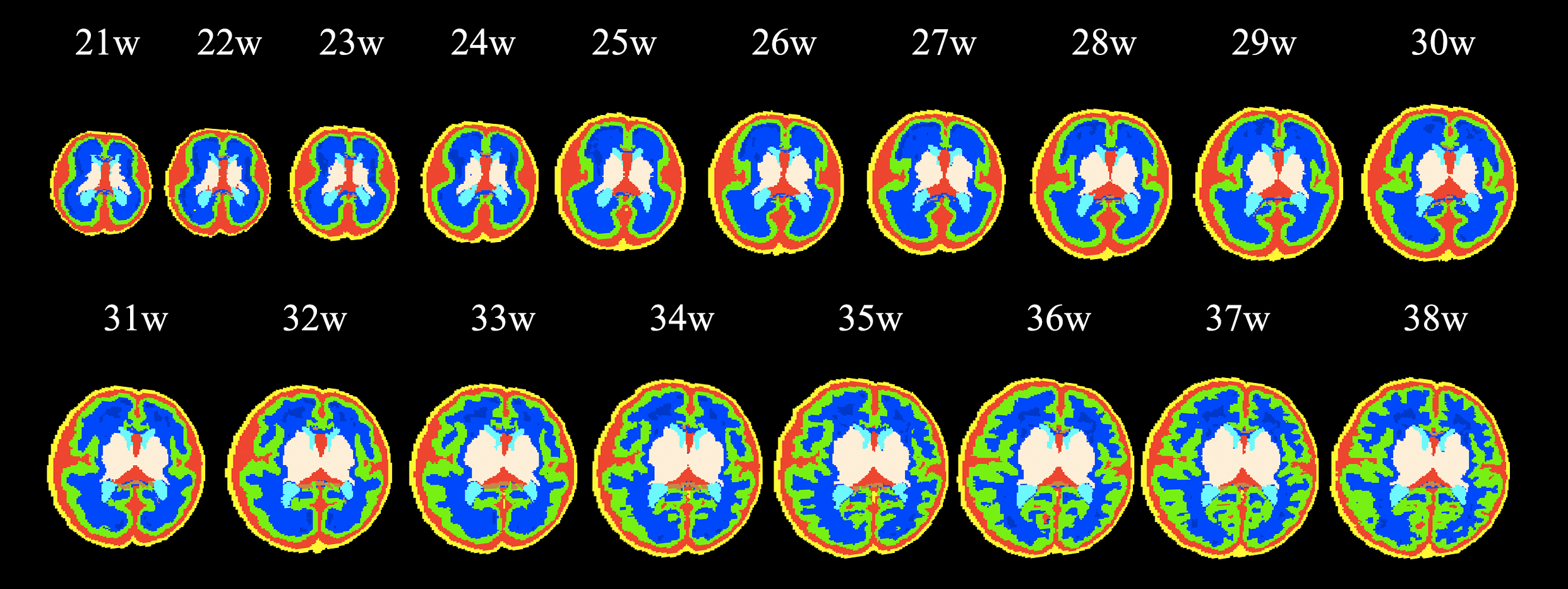}
\caption{Temporal fetal template with the equivalent 7-tissue segmentation.}
\label{fig:FetTemplate}
\end{figure}

Figure \ref{fig:FetTemplate} shows the resulting fetal template segmentation, in which the background label has been added by dilation of the brain mask computed using the BET algorithm.

\subsection{Multi-subject atlas}

A multi-subject fetal atlas has been built selecting 20 subjects and computing on them the same structural segmentation presented in the ALBERTs atlases. In this case, a three-channel registration has been used, including in the registration process the gray matter and ventricle segmentations, alongside the T2-weighted images. Results of the registration are shown in Figure \ref{fig:FetMultiAtlas}:

\begin{figure}[H]
\centering
\includegraphics[width=\linewidth]{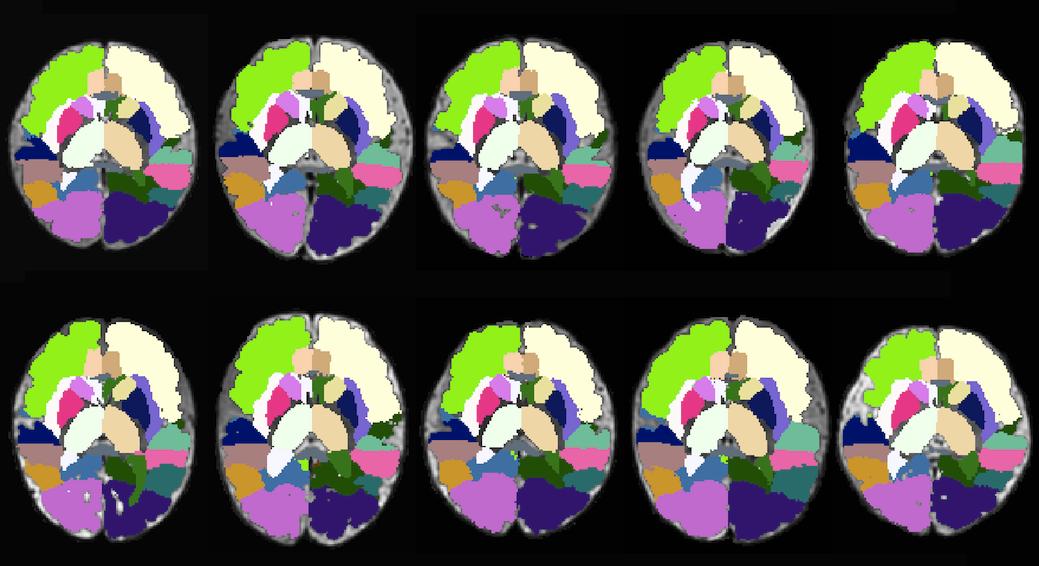}
\caption{Structural segmentation for 10 of the total 20 subjects in the multi-subject atlas.}
\label{fig:FetMultiAtlas}
\end{figure}

\begin{figure}[H]
\centering
  \begin{subfigure}[b]{0.8\textwidth}
    \centering
    \includegraphics[width=\textwidth]{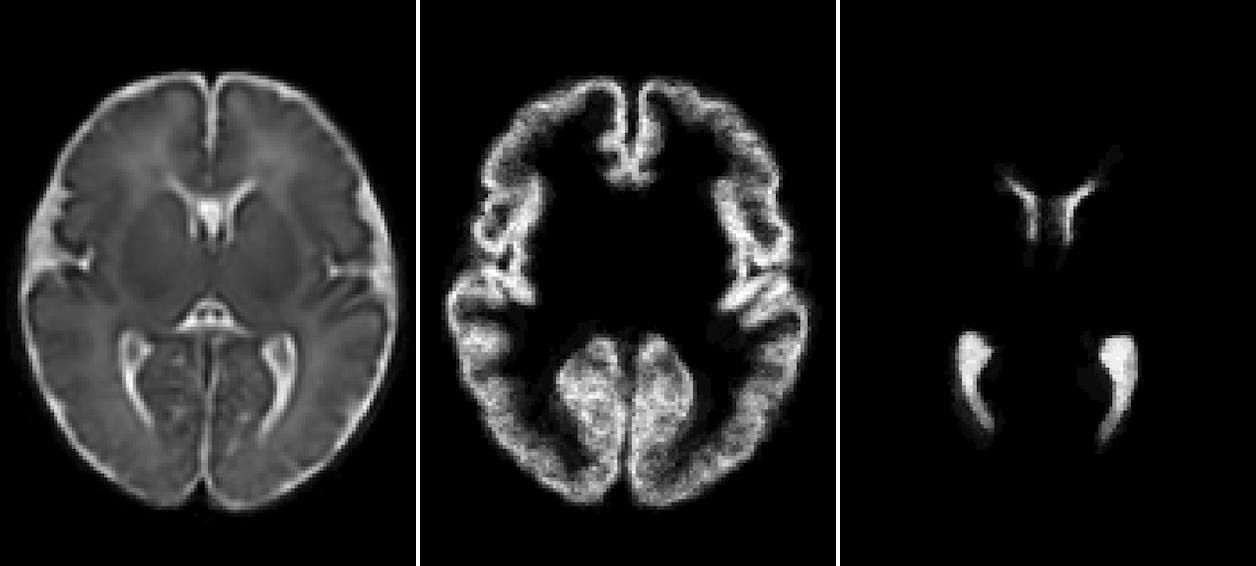}
    \caption{}
    \label{fig:7a}
  \end{subfigure}%
  \vfill
  \begin{subfigure}[b]{0.8\textwidth}
    \centering
    \includegraphics[width=\textwidth]{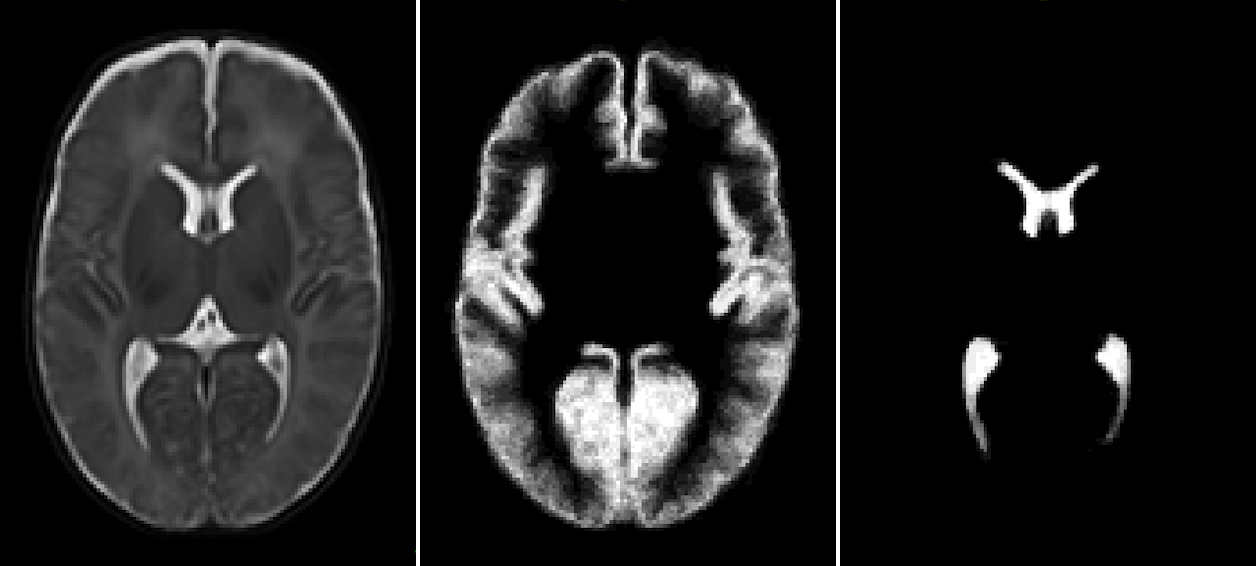}
    \caption{}
    \label{fig:7b}
  \end{subfigure}
\caption{(a) Fetal and (b) neonatal templates obtained from the multi-subject atlases with the cortical gray matter and ventricles probability maps.}
\label{fig:templates}
\end{figure}

In order to perform the multi-subject atlas-based segmentation we had to create first a template for each one of the multi-subject atlases (i.e., fetal and neonatal). The templates resulting from this process are shown in Figure \ref{fig:templates}, with the corresponding probability maps.

\subsection{Segmentation}

\begin{figure}[H]
\centering
  \begin{subfigure}[b]{0.8\textwidth}
    \centering
    \includegraphics[width=\textwidth]{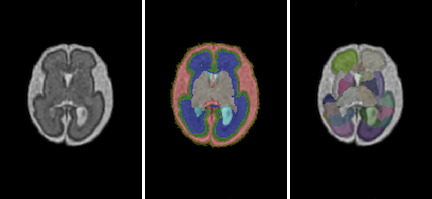}
    \caption{}
    \label{fig:8a}
  \end{subfigure}%
  \vfill
  \begin{subfigure}[b]{0.8\textwidth}
    \centering
    \includegraphics[width=\textwidth]{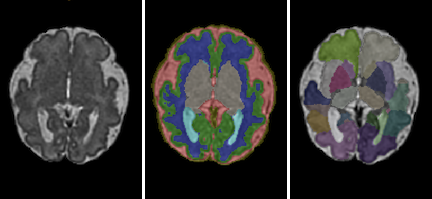}
    \caption{}
    \label{fig:8b}
  \end{subfigure}
\caption{Fetal Segmentation. (a) Early-onset fetal (27 weeks of gestation) T2-weighted image, tissue segmentation and lobular parcellation; (b) Late-onset fetal (32 weeks of gestation) T2-weighted image, tissue segmentation and lobular parcellation.}
\label{fig:Fet}
\end{figure}

Using our pipeline, incorporating the new atlases, we are able to accurately segment both early and late-onset subjects, as shown in Figure \ref{fig:Fet}.

We can analyse how the implemented tools work for tissue and structural segmentations, for acquisitions at different ages, and show how they compare to the only comparable automatic, end-to-end segmentation tool, the dHCP neonatal pipeline.

\begin{figure}[H]
\centering
\includegraphics[width=0.8\linewidth]{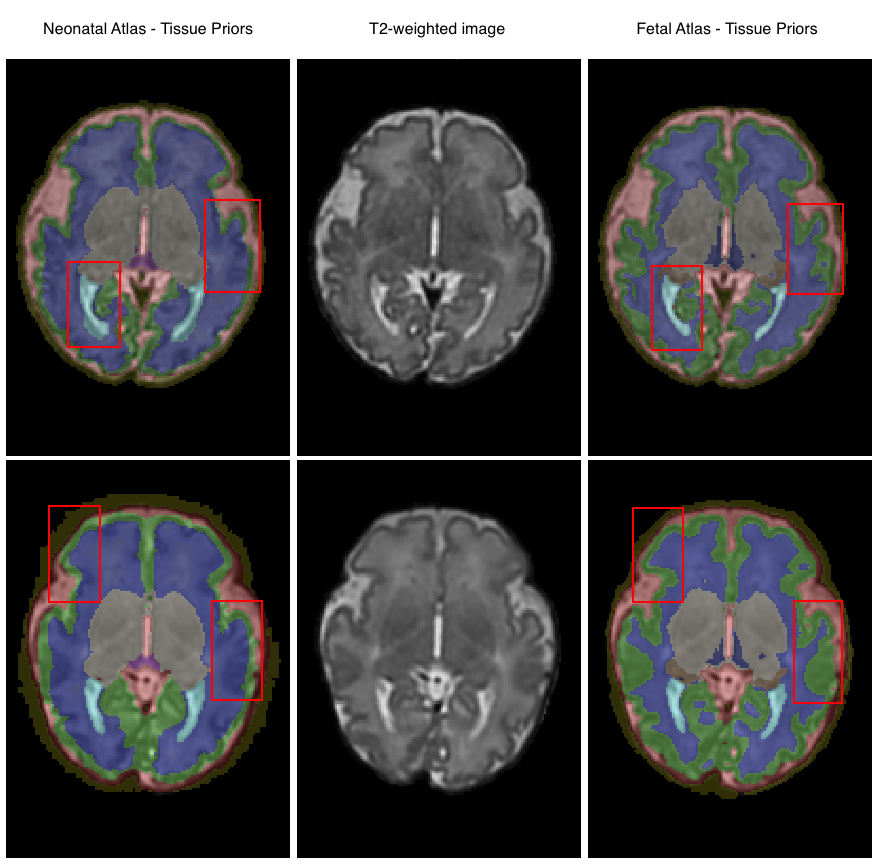}
\caption{Comparison of estimated tissue priors. The figure shows the impact of the use of the enhanced version of the fetal temporal template presented in \cite{630093} on the quality of the tissue priors estimation for a late-onset fetus, with respect to the neonatal temporal template.}
\label{fig:FetVsNeo}
\end{figure}

The initial tissue estimation depends on the temporal template used. The implementation of the modified fetal temporal template in the pipeline leads to visible improvements in the estimation of the cortical gray matter boundaries and of the ventricles, as shown in Figure \ref{fig:FetVsNeo} for a late-onset fetal acquisition segmented both using the dHCP neonatal pipeline and the proposed perinatal pipeline. The estimation of tissue priors is of utmost importance: both gray matter and ventricles priors are used in the multi-subject registration (section \ref{section:Segmentation}), and thus their initial segmentation directly affects the final structural segmentation accuracy.

\begin{figure}[H]
\centering
  \begin{subfigure}[b]{\textwidth}
    \centering
    \includegraphics[width=\textwidth]{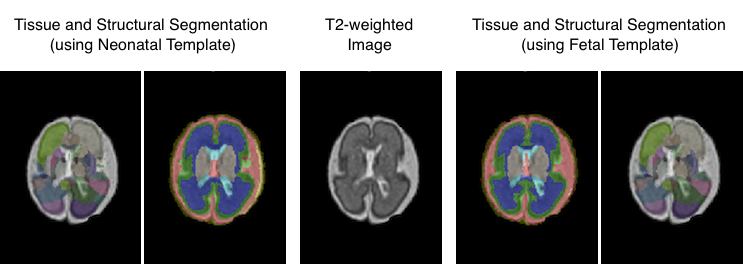}
    \caption{}
    \label{fig:10a}
  \end{subfigure}%
  \vfill
  \begin{subfigure}[b]{\textwidth}
    \centering
    \includegraphics[width=\textwidth]{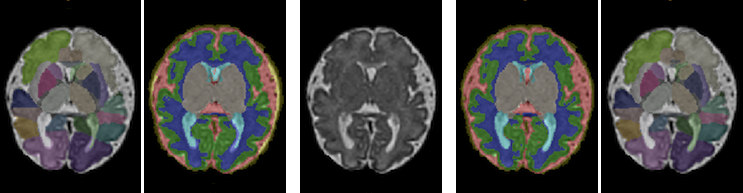}
    \caption{}
    \label{fig:10b}
  \end{subfigure}
  \vfill
  \begin{subfigure}[b]{\textwidth}
    \centering
    \includegraphics[width=\textwidth]{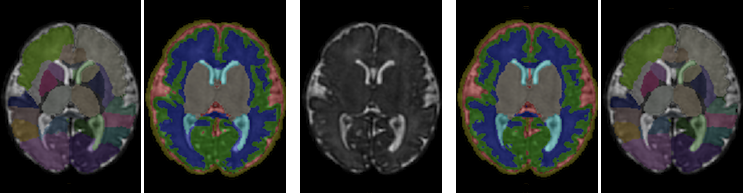}
    \caption{}
    \label{fig:10b}
  \end{subfigure}
\caption{Segmentation accuracy qualitative comparison. The figure shows a comparison between the segmentations of 3 fetuses at different gestational ages. (a) Tissue and lobular segmentation at 26.3 weeks of gestation, using a neonatal template and a fetal template respectively. (b) Tissue and lobular segmentation at 33.9 weeks of gestation. (c) Tissue and lobular segmentation at 37.5 weeks of gestation.}
\label{fig:Comparison}
\end{figure}

By using the newly developed fetal templates and modifying the registration method, we can see that the modified pipeline adapts better to early and late-onset fetal acquisitions.  Figure \ref{fig:Comparison} shows how the use of a fetal structural multi-subject atlas improves the segmentation of the cortical plate, the CSF and the white matter compared to using the nenonatal pipeline. 

Quantitative results are given for a subset of 70 fetal acquisitions (Table \ref{fig:TableDice}). Ground truth for these subjects were provided by manual correction of an expert and compared to the segmentations obtained using the original dHCP neonatal pipeline and our perinatal pipeline. It can be seen that the proposed pipeline improves the segmentation results for cGM, white matter and ventricles. 

Moreover, the registration methodology developed allows reducing dramatically the computational load for each segmentation. The time for a single segmentation went from over 3 hours to roughly 15 minutes on an 8-core machine. More importantly, the number of pairwise registrations is made constant and equal to one, regardless of the size of the multi-subject atlases. 

\begin{table}
\caption{Segmentation accuracy. The table reports the dice coefficient, with respect to the ground truth provided by an expert, of the resulting tissue segmentations using the dHCP pipeline and the proposed perinatal pipeline. The mean dice coefficient (with $\pm$ standard deviation) is reported for each segmented tissue.}
\centering
\begin{tabular}[\linewidth]{*3r}
 Tissue & Neonatal Atlas & Fetal Atlas \\
\midrule
 CSF & 0.79 $\pm$ 0.02 & 0.83 $\pm$ 0.04 \\ 
 Cortical Plate & 0.76 $\pm$ 0.05 & 0.85 $\pm$ 0.03 \\
 White Matter & 0.85 $\pm$ 0.02 & 0.90 $\pm$ 0.02 \\
 Ventricles & 0.66 $\pm$ 0.07 & 0.73 $\pm$ 0.05 \\
 Cerebellum & 0.89 $\pm$ 0.03 & 0.93 $\pm$ 0.02 \\
 Brainstem & 0.88 $\pm$ 0.03 & 0.92 $\pm$ 0.02 \\
 Deep Gray Matter & 0.90 $\pm$ 0.02 & 0.91 $\pm$ 0.01 \\
\end{tabular}
\label{fig:TableDice}
\end{table}

The three-channel registration has proven to guarantee more accurate results, compared to the simple T2 intensity-based registration and a two-channel registration using the gray matter probability maps. Tissue segmentation can be derived from the structural segmentation and compared to the ground truth provided by an expert. Dice coefficient was computed to quantify the overlap with the ground truth segmentation. As shown in Table \ref{tab:DCreg}, segmentation improves for the internal structures (e.g.: ventricles, deep gray matter). 

\begin{table}[H]
\caption{Registration method comparison. The table reports the dice coefficient, with respect to the ground truth provided by an expert, of the resulting tissue segmentations using one, two and three channels for the registration respectively. The mean dice coefficient (with $\pm$ standard deviation) is reported for each tissue segmented.}
\centering
\begin{tabular}[\linewidth]{*4r}
 Tissue & T2-based & 2-channel & 3-channel \\ 
 \midrule
 CSF & 0.83 $\pm$ 0.02 & 0.85 $\pm$ 0.01 & 0.83 $\pm$ 0.04 \\ 
 Cortical Plate & 0.75 $\pm$ 0.05 & 0.83 $\pm$ 0.03 & 0.85 $\pm$ 0.03 \\
 White Matter & 0.85 $\pm$ 0.02 & 0.89 $\pm$ 0.02 & 0.90 $\pm$ 0.02 \\
 Ventricles & 0.65 $\pm$ 0.07 & 0.68 $\pm$ 0.05 & 0.73 $\pm$ 0.05 \\
 Cerebellum & 0.92 $\pm$ 0.03 & 0.91 $\pm$ 0.03 & 0.93 $\pm$ 0.02 \\
 Brainstem & 0.91 $\pm$ 0.03 & 0.88 $\pm$ 0.03 & 0.92 $\pm$ 0.02 \\
 Deep Gray Matter & 0.90 $\pm$ 0.02 & 0.90 $\pm$ 0.02 & 0.91 $\pm$ 0.01
\end{tabular}
\label{tab:DCreg}
\end{table}

\subsection{Surface extraction}

\begin{figure}[h]
  \begin{subfigure}{0.3\textwidth}
    \centering
    \includegraphics[width=\textwidth]{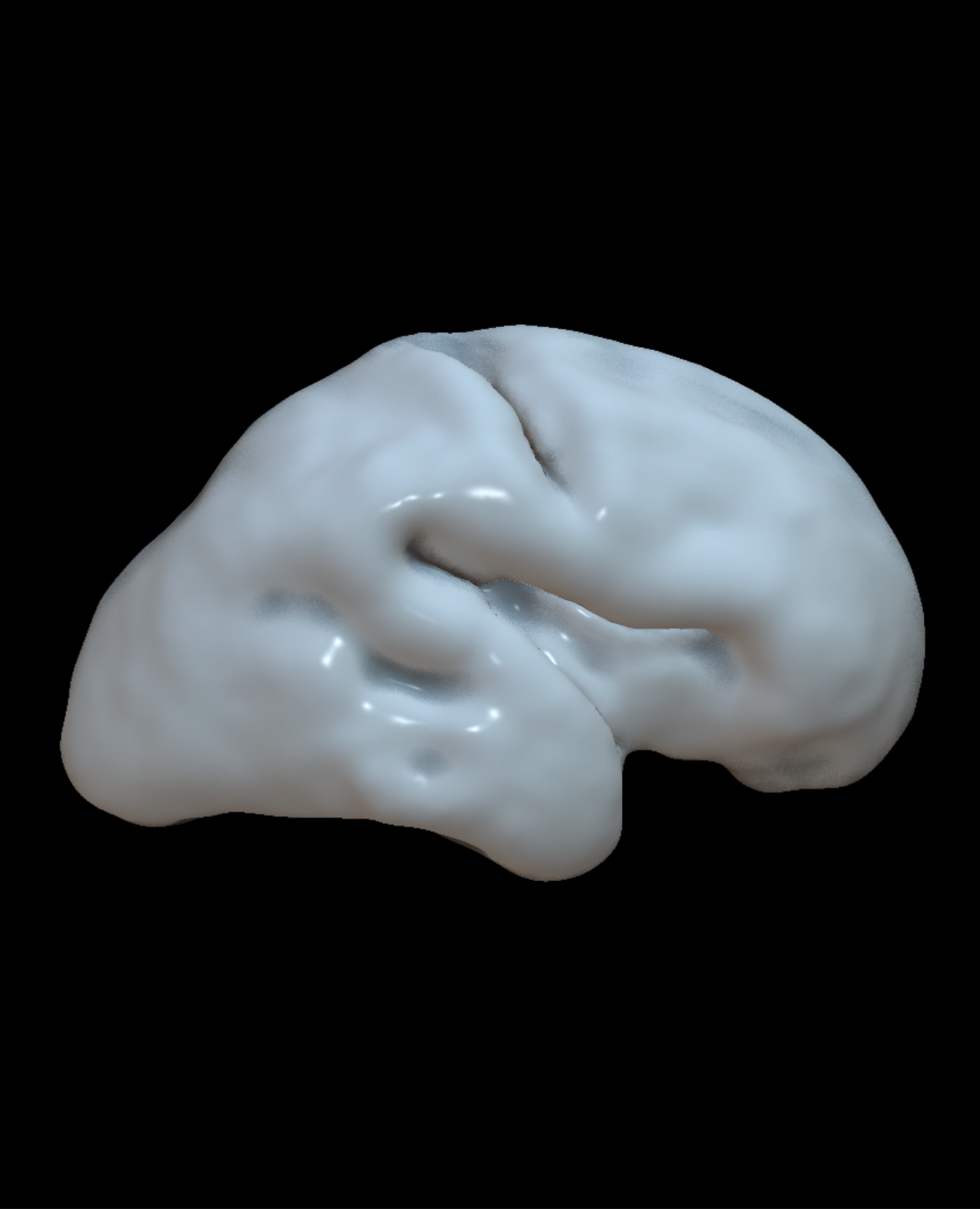}
    \caption{}
    \label{fig:11a}
  \end{subfigure}%
  \hfill
  \begin{subfigure}{0.3\textwidth}
    \centering
    \includegraphics[width=\textwidth]{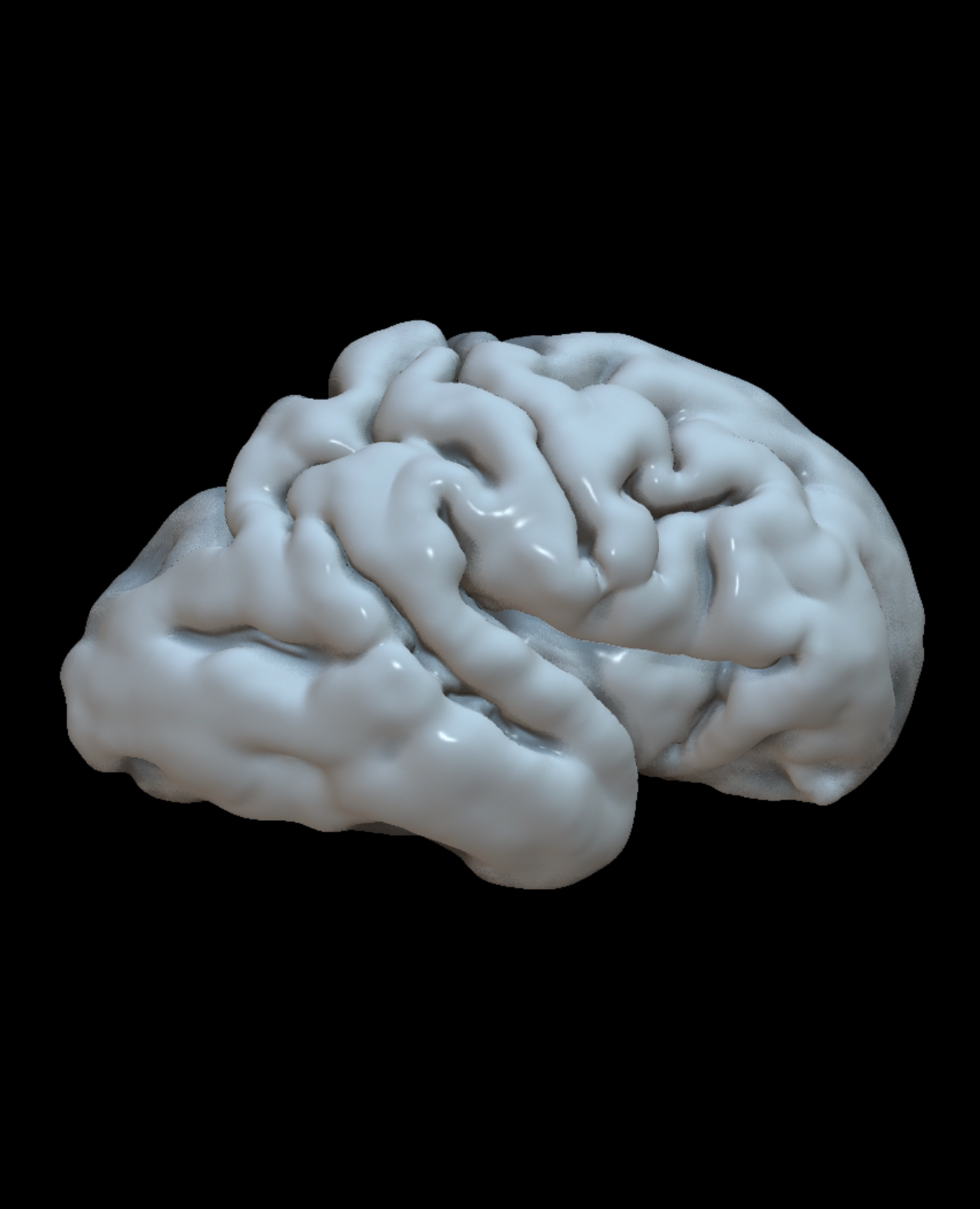}
    \caption{}
    \label{fig:11b}
  \end{subfigure}
  \hfill
  \begin{subfigure}{0.3\textwidth}
    \centering
    \includegraphics[width=\textwidth]{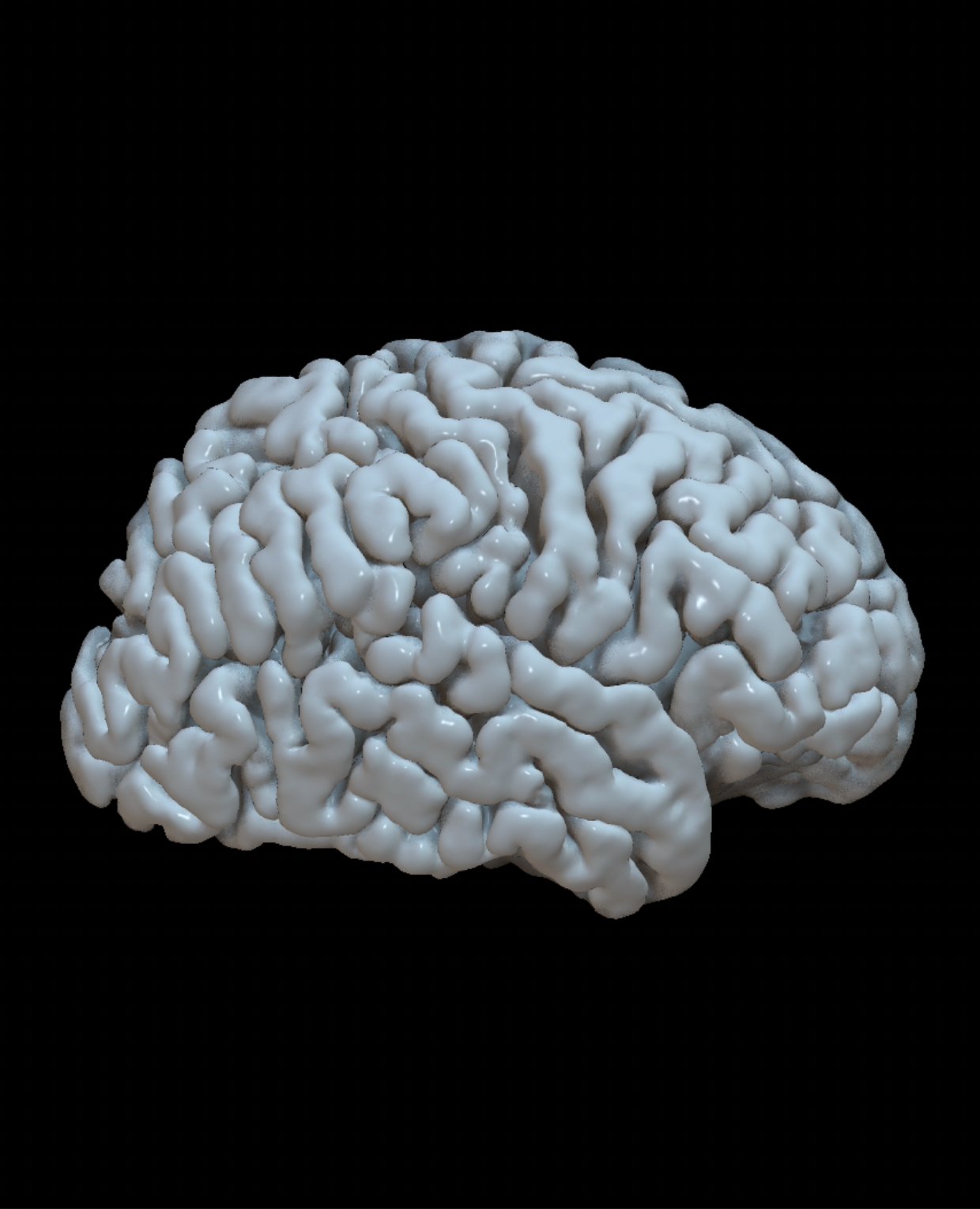}
    \caption{}
    \label{fig:11c}
  \end{subfigure}
\caption{Extracted surfaces for fetuses and neonates. The figure shows the increasing complexity and convoluted shape of the cortex for (a) an early-onset fetus (26.7 weeks of gestation), (b) a late-onset fetus (32.4 weeks of gestation) and (c) a neonate (44.1 weeks of gestation).}
\label{fig:NeoFetRecon}
\end{figure}

Extraction of the cortical surface from segmentation allows analyzing differences in shape, and thus gives more information on the brain development. Reconstruction can be applied to fetal and neonatal subjects using the enhanced pipeline. Figure \ref{fig:NeoFetRecon} shows the reconstruction for an early stage fetus, for a late stage fetus and a neonate. 

\subsection{Feature Extraction}

After the cortical surface extraction, shape features of interest for the brain development assessment can be computed using the calculated meshes. In particular, we focus on curvature, thickness, sulcal depth and local gyrification index. 

\subsubsection{Curvature}

Surface mean curvature is estimated from the average of the principal curvatures of the white matter surface. The principal curvatures represent the minimum and maximum bending of a regular surface at each given point. Mean curvature tends to increase while the cortex grows and gets more convoluted, whereas at an early stage in the development, the less complex shape of the cortical plate involves a lower mean curvature, as it can be seen in Figure \ref{fig:12a}.

\subsubsection{Local Gyrification Index}

Local Gyrification Index (LGI) is computed as the ratio between an area taken on the cortical surface of the brain and the area covered by the same points on the inflated surface. The inflation process moves outwards the points in the sulci and brings in the points on the gyral crowns - preserving distance between neighboring vertices - until a certain smoothness is reached. It can be seen on the average fetal brain as LGI tends to zero where the surface is flat, while it has higher values in the depth of the main sulci (Figure \ref{fig:12b}).

\subsubsection{Sulcal Depth}

\begin{figure}[H]
  \begin{subfigure}[b]{\textwidth}
    \centering
    \includegraphics[width=\textwidth]{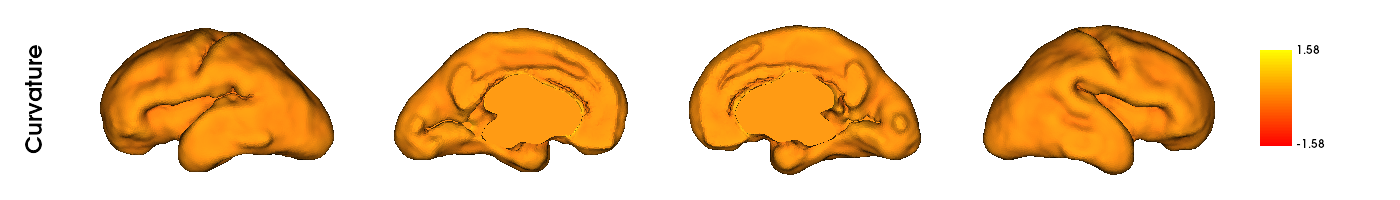}
    \caption{}
    \label{fig:12a}
  \end{subfigure}%
  \vfill
  \begin{subfigure}[b]{\textwidth}
    \centering
    \includegraphics[width=\textwidth]{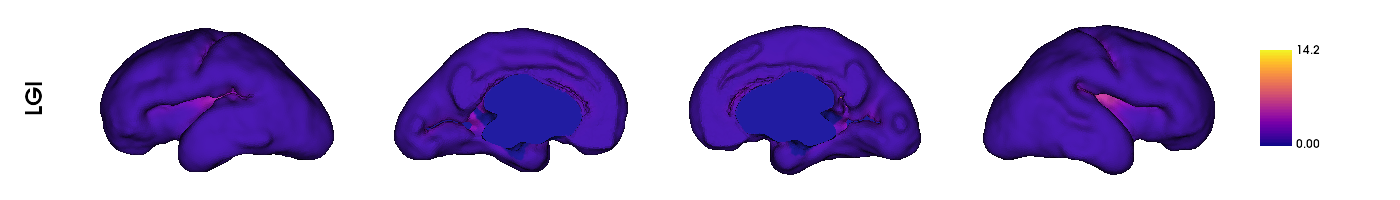}
    \caption{}
    \label{fig:12b}
  \end{subfigure}
  \vfill
  \begin{subfigure}[b]{\textwidth}
    \centering
    \includegraphics[width=\textwidth]{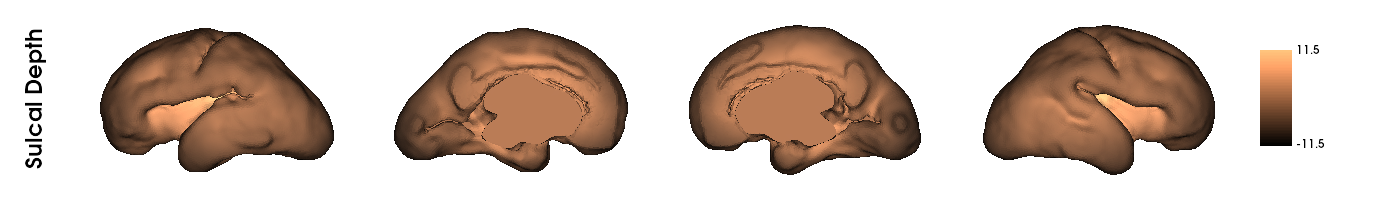}
    \caption{}
    \label{fig:12c}
  \end{subfigure}
  \vfill
  \begin{subfigure}[b]{\textwidth}
    \centering
    \includegraphics[width=\textwidth]{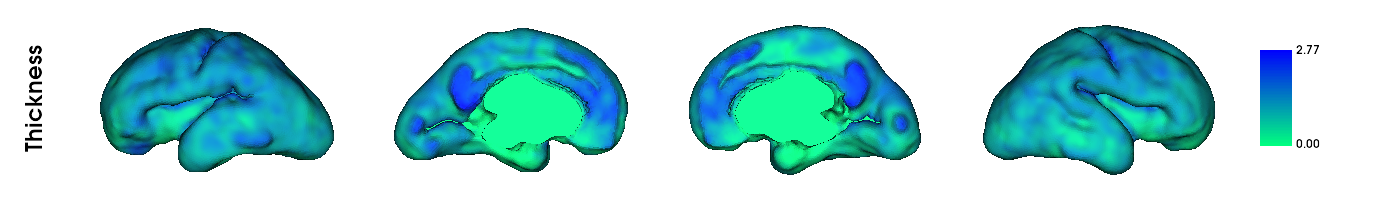}
    \caption{}
    \label{fig:12d}
  \end{subfigure}
\caption{Cortical morphological features of a fetal subject. The figure shows the extracted cortical feature for a fetus, in particular (a) mean curvature, (b) local gyrification index (LGI), (c) sulcal depth and (d) thickness.}
\label{fig:Features}
\end{figure}

Sulcal depth represents the average convexity or concavity of cortical surface points. Being directly related to the presence of sulci on the cortex, it is higher in the more convoluted parts of the brain (Figure \ref{fig:12c}) and it naturally increases while the brain grows, whereas for a fetal brain has lower values.

\subsubsection{Thickness}

Cortical thickness is defined as the average distance between two measures:
1) the Euclidean distance from the white surface to the closest vertex in the pial surface;
2) the Euclidean distance from the pial surface to the closest vertex in the white surface.

Thickness is higher at early stage, as the cortex becomes thinner to be able to bend on itself while it grows.

\section{Discussion}
\label{ch:Discussion}

This paper presents a new pipeline, built upon the structure of the dHCP neonatal pipeline \cite{makro2018}, for the structural segmentation and cortical surface extraction of fetal and neonatal brain MRI. The modification and adaptation of an existing fetal temporal template, and the creation of a new multi-subject fetal atlas are also reported. This pipeline allows for a consistent and coherent fetal and neonatal segmentation in the same framework, for the analysis of perinatal data and thus with potential application to longitudinally acquired data. Moreover, a new registration approach has been proposed and, consequently, the segmentation has been made more robust by adding a new channel (i.e. the ventricle probability maps) to the multi-subject atlas registration.

The modifications applied to the registration process lead to an improved neonatal segmentation compared to the dHCP neonatal pipeline. When compared on a fetal dataset, results show how the segmentation accuracy was enhanced by using fetal templates for all the main tissues: considering both the external cortical surface and the internal shape of the ventricles in the transformation leads to a more accurate overall registration. 

Moreover, for both neonatal and fetal acquisitions, the proposed registration approach leads to a reduced computational load, and it makes the process scalable. This leads to the possibility of increasing the number of the subjects in both the fetal and neonatal atlases, enhancing their validity without affecting the computational load of the pipeline.

Finally, the surface extraction process has proven to be efficient for fetal images, based on an expert qualitative analysis, faithfully reporting the convoluted shape of the cortical surface resulting from the segmentation in the 3D regularised mesh computation. The possibility of computing a structural and regional segmentation for both fetal and neonatal images, and to extract the cortical surfaces, opens a variety of possible analysis to be carried out in the future. 

Future works will focus on further improving and enhancing the performance of the pipeline, with the possible implementation of deep learning-based approaches for registration, which would further reduce the computational load and shorten the segmentation process. It would also be possible to increase the size of the used multi-subject atlases, including also pathological anatomies and widening the temporal window covered by the subjects to enhance the range of application of the pipeline. Furthermore, of particular interest from a clinical perspective could be the analysis of local cortical features in longitudinally acquired data, and their relationship with other brain features and diseases.

\section{Acknowledgements}
\label{ch:Acknowledgements}

We thank the study participants for their personal time and commitment to this research. We also thank all the medical staff, technicians, midwives and nurses of BCNatal, especially Giulia Casu and Killian Vellvé for participants recruitment; Mònica Rebollo, Laura Oleaga and Cesar Garrido for MRI assessment.

This publication is part of the project PCI2021-122044-2A, funded by the project ERA-NET NEURON Cofund2, by MCIN/AEI/10.13039/501100011033/ and by the European Union “NextGenerationEU”/PRTR. The project that gave rise to these results received funding from the European Union's Horizon 2020 research and innovation programme under the Marie Sklodowska-Curie grant agreement No. 713673 from the Instituto de Salud Carlos III (PI18/00073; PI20/00246; INT21/00027) within the Plan Nacional de I+D+I and cofinanced by ISCIII-Subdirección General de Evaluación together with the Fondo Europeo de Desarrollo Regional (FEDER) “Una manera de hacer Europa”, the Cerebra Foundation for the Brain Injured Child (Carmarthen, Wales, UK) and ASISA foundation. A. Urru has received the support of a fellowship from ”la Caixa” Foundation under grant nº LCF/BQ/DI17/11620069. Additionally, A. Nakaki has received the support of a fellowship from ”la Caixa” Foundation under grant nº LCF/BQ/DR19/11740018 and E. Eixarch has received funding from the Departament de Salut under grant SLT008/18/00156. G. Piella was supported by ICREA under the ICREA Academia programme.

\bibliography{mybibfile}

\end{document}